\documentclass[12pt,a4paper]{article}
\usepackage{ecsia}

\usepackage{graphicx}

\usepackage{amsmath}
\usepackage{amsfonts}
\usepackage{amssymb}

\usepackage{mathrsfs}
\usepackage{scalerel}

\newcommand\scale[2]{\vstretch{#1}{\hstretch{#1}{#2}}}
\newcommand{\LIPplus}{\mathbin{\ooalign{$\bigtriangleup$\crcr\hidewidth
  \raise.14em\hbox{$\scale{0.7}{\scriptscriptstyle+}$}\hidewidth}}}
\newcommand{\LIPminus}{\mathbin{\ooalign{$\bigtriangleup$\crcr\hidewidth
  \raise.14em\hbox{$\scale{0.7}{\scriptscriptstyle-}$}\hidewidth}}}
\newcommand{\LIPtimes}{\mathbin{  \ooalign{$\bigtriangleup$\crcr\hidewidth
  \raise.14em\hbox{$\scale{0.7}{\scriptscriptstyle\times}$}\hidewidth}}}

\newcommand{\Real}{\mathbb R}

\newcommand{\la}{\lambda}

\newcommand{\I}{\mathcal{I}}

\usepackage{verbatim}

\DeclareFontFamily{U}{mathx}{\hyphenchar\font45}
\DeclareFontShape{U}{mathx}{m}{n}{
      <5> <6> <7> <8> <9> <10>
      <10.95> <12> <14.4> <17.28> <20.74> <24.88>
      mathx10
      }{}
\DeclareSymbolFont{mathx}{U}{mathx}{m}{n}
\DeclareFontSubstitution{U}{mathx}{m}{n}
\DeclareMathAccent{\widecheck}{0}{mathx}{"71}

\begin{document}

\parbox[c]{0.925\textwidth}{

\title{A simple expression for the map of Asplund's distances with the multiplicative Logarithmic Image Processing (LIP) law}

\author{Guillaume Noyel\corr, and Michel Jourlin}

\affil{International Prevention Research Institute, 95 cours Lafayette,	69006 Lyon, France, www.i-pri.org}
}

\cite{Noyel2017} have shown that the map of Asplund's distances, using the multiplicative LIP law, between an image, $f \in \overline{\I}=[0,M]^{D}$, $D \subset \Real^N$, $M \in \Real$, $M > 0$, and a structuring function $B \in [O,M]^{D_B}$, $D_B \subset \Real^N$, namely the probe, is the logarithm of the ratio between a general morphological dilation $\la_{B} f$ and an erosion $\mu_{B} f$, in the lattice $(\overline{\I} , \leq)$:
\begin{equation}
	As_{B}^{\LIPtimes}f = \ln \left( \frac{ \la_{B} f }{ \mu_{B} f }\right) = 
	\ln{ \left( \frac{ \vee_{h \in D_B} \{ \widetilde{f}(x+h) / \widetilde{B}(h) \} }{ \wedge_{h \in D_B} \{ \widetilde{f}(x+h) / \widetilde{B}(h) \} }\right) }, \> \text{ with } f > 0 \text{ and } \widetilde{f}= \ln{\left( 1 - f/M \right)}.
	\label{eq:map_As_la_mu_general_se}
\end{equation}

The morphological dilation and erosion with a translation invariance (in space and in grey-level) are defined, for additive structuring functions as  \citep{Serra1988,Heijmans1990}:
\begin{equation}
\begin{array}{ccc}
(\delta_B(f))(x) 			&=& \vee_{h \in D_B} 	 \left\{ f(x - h) + B(h) \right\} = (f \oplus B) (x)  \> \text{, (dilation)}\\
(\varepsilon_B(f))(x) &=& \wedge_{h \in D_B} \left\{ f(x + h) - B(h) \right\} = (f \ominus B) (x) \> \text{, (erosion)}\\
\end{array}
\label{eq:erode_dilate_funct_add}
\end{equation}
and for multiplicative structuring functions as \citep{Heijmans1990}:
\begin{equation}
\begin{array}{ccl}
\vee_{h \in D_B} 	 \left\{ f(x - h) . B(h) \right\} &=& (f \dot{\oplus} B) (x)  \> \text{, (dilation)}\\
\wedge_{h \in D_B} \left\{ f(x + h) / B(h) \right\} &=& (f \dot{\ominus} B) (x) \> \text{, (erosion)}.\\
\end{array}
\label{eq:erode_dilate_funct_mult}
\end{equation}
If $f>0$ and $B>0$, we have $(f \dot{\oplus} B) = \exp{( \vee_{h \in D_B} \left\{ \ln{f(x-h)} + \ln{B(h)} ) \right\})} = \exp{ (\ln f \oplus \ln B )}$.
Let us define, the reflected structuring function by $\overline{B}(x)=B(-x)$ 
and $\widehat{f} = \ln{(- \widetilde{f})}$. We obtain:
\begin{equation}
\begin{array}{cll}
\la_{B} f &=& \vee_{h \in D_B} \left\{ \widetilde{f}(x+h) / \widetilde{B}(h) \right\} = \vee_{-h \in D_B} \left\{ \widetilde{f}(x-h) / \widetilde{B}(-h) \right\}= (-\widetilde{f}) \dot{\oplus} (-1/\widetilde{\overline{B}})\\
&=& \exp{ \left( \vee_{-h \in D_B} \left\{ \ln{(-\widetilde{f}(x-h))} + \ln{(-1/\widetilde{\overline{B}}(h)) } \right\}   \right) } \> \text{, with } \widetilde{f}<0 \text{ and } \widetilde{\overline{B}} <0\\
&=& \exp{ \left( \ln{(-\widetilde{f})} \oplus (- \ln{(-\widetilde{\overline{B}})) } \right) } =  \exp{ \left( \widehat{f} \oplus (- \widehat{\overline{B}}) \right)}.
\end{array}
\label{eq:dem:la}
\end{equation}
Similarly, we have $\mu_{B} f = (-\widetilde{f}) \dot{\ominus} (-\widetilde{B}) = \exp{ \left( \widehat{f} \ominus \widehat{B} \right) } $.
Using the previous expressions of $\la_{B} f$ and $\mu_{B} f$ into equation \ref{eq:map_As_la_mu_general_se}, we obtain:
\begin{equation}
	As_{B}^{\LIPtimes}f = \ln \left( \frac{ \la_{B} f }{ \mu_{B} f }\right) 
	= \ln \left( \frac{ \exp{ \left( \widehat{f} \oplus (- \widehat{\overline{B}}) \right) } } { \exp{ \left( \widehat{f} \ominus \widehat{B} \right)} }\right)
	= \left[ \widehat{f} \oplus (- \widehat{\overline{B}}) \right] - \left[ \widehat{f} \ominus \widehat{B} \right]
	= \delta_{-\widehat{\overline{B}}} \widehat{f} - \varepsilon_{\widehat{B}} \widehat{f}.
	\label{eq:map_As_la_mu_general_se_simple}
\end{equation}
Therefore, the map of Asplund's distances with the LIP multiplication is the difference between a dilation and an erosion with an additive structuring function (i.e. a morphological gradient). 

\bibliographystyle{iasart} 
\bibliography{refs}

\end{document}